\documentclass[10pt,twocolumn,letterpaper]{article}

\usepackage{cvpr}
\usepackage{times}
\usepackage{epsfig}
\usepackage{graphicx}
\usepackage{amsmath}
\usepackage{amssymb}
\usepackage{algorithm}
\usepackage{enumitem}
\usepackage{algpseudocode}
\usepackage{subcaption}
\usepackage{wrapfig}
\usepackage[font=footnotesize]{caption}
\usepackage{makecell}

\newcommand{\x}{{\bf x}}
\newcommand{\w}{{\bf w}}
\newcommand{\W}{{\bf \mathsf{W}}}

\newtheorem{theor}{Theorem}
\newtheorem{lemma}{Lemma}
\newtheorem{corollary}{Corollary}

\usepackage[pagebackref=true,breaklinks=true,letterpaper=true,colorlinks,bookmarks=false]{hyperref}

\cvprfinalcopy 
\setlist[itemize]{leftmargin=4.mm}

\def\cvprPaperID{2650} 

\ifcvprfinal\fi
\begin{document}

\title{Wasserstein Introspective Neural Networks}

\newcommand*\samethanks[1][\value{footnote}]{\footnotemark[#1]}
\author{Kwonjoon Lee\qquad Weijian Xu\qquad Fan Fan\qquad Zhuowen Tu\\
University of California San Diego\\
{\tt\small $\{$kwl042, wex041, f1fan, ztu$\}$@ucsd.edu}
}

\maketitle
\thispagestyle{empty}

\begin{abstract}
   We present Wasserstein introspective neural networks (WINN) that are both a generator and a discriminator within a single model. WINN provides a significant improvement over the recent introspective neural networks (INN) method by enhancing INN's generative modeling capability. WINN has three interesting properties: (1) A mathematical connection between the formulation of the INN algorithm and that of Wasserstein generative adversarial networks (WGAN) is made. (2) The explicit adoption of the Wasserstein distance into INN results in a large enhancement to INN, achieving compelling results even with a single classifier --- e.g., providing nearly a 20 times reduction in model size over INN for unsupervised generative modeling. (3) When applied to supervised classification, WINN also gives rise to improved robustness against adversarial examples in terms of the error reduction.
   In the experiments, we report encouraging results on unsupervised learning problems including texture, face, and object modeling, as well as a supervised classification task against adversarial attacks.
    Our code is available online\footnote{\url{https://github.com/kjunelee/WINN}}.
\end{abstract}

\section{Introduction}

\begin{figure}[!htp]
\vspace{-1mm}
\begin{center}
\begin{tabular} {c}
\hspace{-5mm} \includegraphics[width=0.47\textwidth,height=0.41\textwidth ]{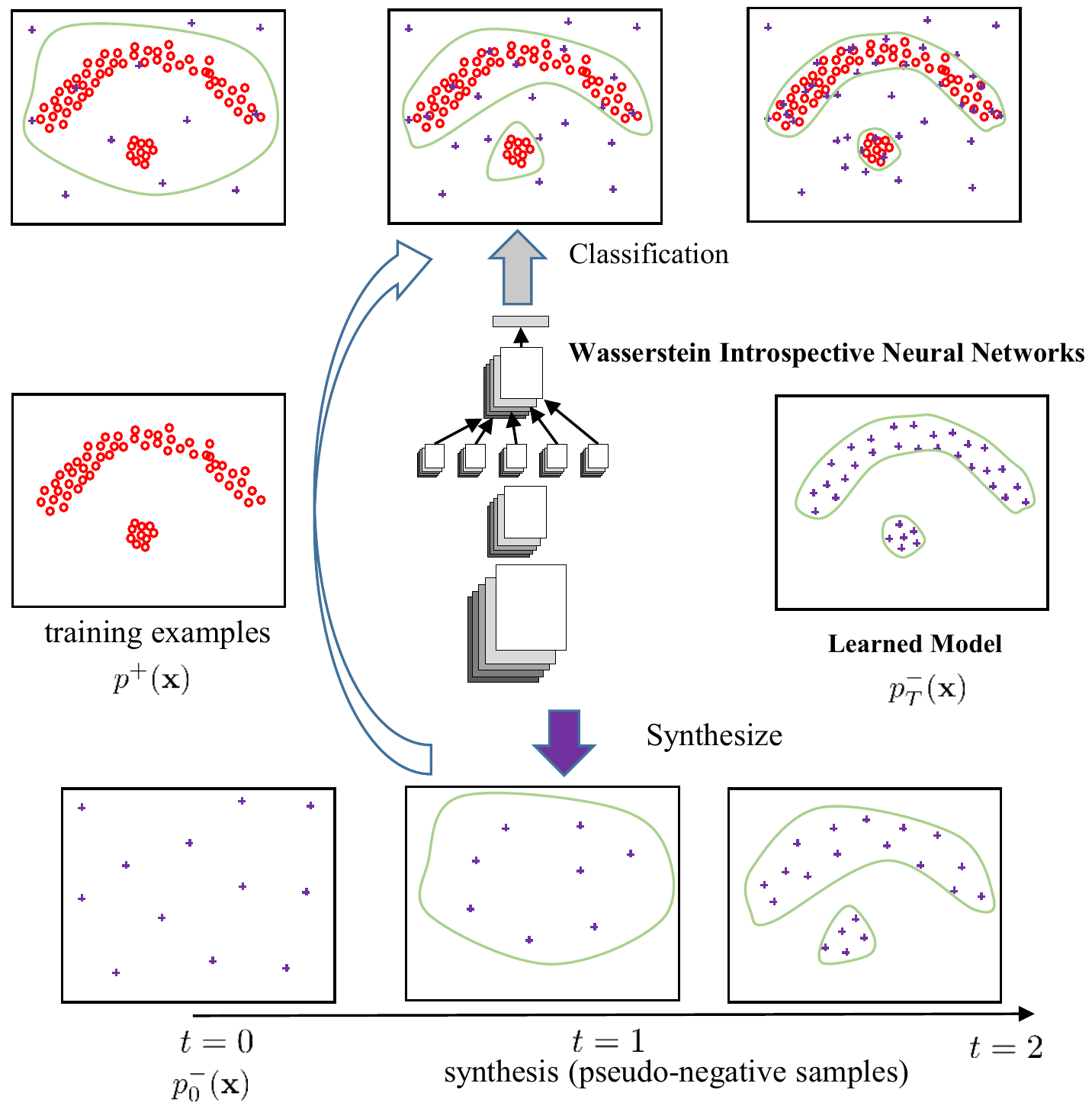} 
\end{tabular}
\end{center}
\vspace{-3mm}
\caption{\footnotesize Schematic illustration of Wasserstein introspective neural networks for unsupervised learning. The left figure shows the input examples; the bottom figures show the pseudo-negatives (purple crosses) being progressively synthesized; the top figures show the classification between the given examples (positives) and synthesized pseudo-negatives (negatives). The right figure shows the model learned to approach the target distribution based on the given data.}
\vspace{-3mm}
\label{fig:INN_pipeline}
\end{figure}

Performance within the task of supervised image classification has been vastly improved in the era of deep learning using modern convolutional neural network (CNN) \cite{CNN} based discriminative classifiers \cite{krizhevsky2012imagenet,szegedy2015going,DSN,simonyan2015very,he2016deep,xie2017aggregated,huang2017dense}. On the other hand, unsupervised generative models in deep learning were previously attained using methods under the umbrella of graphical models --- e.g., the Boltzmann machine \cite{hinton2006fast} or autoencoder \cite{baldi2012autoencoders,kingma2014auto} architectures. However, the rich representational power seen within convolution-based (discriminative) models is not being directly enjoyed in these generative models. Later, inverting convolutional neural networks in order to convert internal representations into a real image was investigated in \cite{dosovitskiy2015learning,kulkarni2015deep}. Recently, generative adversarial networks (GAN) \cite{goodfellow2014generative} and followup works \cite{radford2015unsupervised,WGAN,WGAN-GP} have attracted a tremendous amount of attention in machine learning and computer vision by producing high quality synthesized images by training a pair of competing models against one another in an adversarial manner. While a generator tries to create ``fake'' images to fool the discriminator, the discriminator attempts to discern between these ``real'' (given training) and ``fake'' images. After convergence, the generator is able to produce images faithful to the underlying data distribution.

Before the deep learning era \cite{hinton2006fast}, generative modeling had been an area with a steady pace of development \cite{ackley1985learning,della1997inducing,zhu1997minimax,olshausen1997sparse,wu2010learning,zhu2007stochastic}. These models were guided by rigorous statistical theories which, although nice in theory, did not succeed in producing synthesized images with practical quality.

In terms of building generative models from discriminative classifiers, there have been early attempts in \cite{welling2002self,tu2007learning}. In \cite{welling2002self}, a generative model was obtained from a repeatedly trained boosting algorithm \cite{freund1997decision} using a weak classifier whereas \cite{tu2007learning} used a strong classifier in order to self-generate negative examples or ``pseudo-negatives''.

To address the lack of richness in representation and efficiency in synthesis, convolutional neural networks were adopted in introspective neural networks (INN) \cite{Lazarow2017intro,Jin2017intro} to build a single model that is simultaneously generative and discriminative. The generative modeling aspect was studied in \cite{Lazarow2017intro} where a sequence of CNN classifiers ($10-60$) were trained, while the power within the classification setting was revealed in \cite{Jin2017intro} in the form of introspective convolutional networks (ICN) that used only a single CNN classifier. Although INN models \cite{Lazarow2017intro,Jin2017intro} point to a promising direction to obtain a single model being both a good generator and a strong discriminative classifier, a sequence of CNNs were needed to generate realistic synthesis. As a result, this requirement may serve as a possible bottleneck with respect to training complexity and model size.

Recently, a generic formulation \cite{WGAN} was developed within the GAN model family to incorporate a Wasserstein objective to alleviate the well-known difficulty in GAN training. Motivated by introspective neural networks (INN) \cite{Lazarow2017intro} and this Wasserstein objective \cite{WGAN}, we propose to adopt the Wasserstein term into the INN formulation to enhance the modeling capability. The resulting model, Wasserstein introspective neural networks (WINN) shows greatly enhanced modeling capability over INN by having $20\times$ reduction in the number of CNN classifiers. 

\section{Significance and Related Work}
We make several interesting observations for WINN:
{\small 
\begin{itemize}
 \setlength\itemsep{0mm}
 \setlength{\itemindent}{0mm}
\item A mathematical connection between the WGAN formulation \cite{WGAN} and the INN algorithm \cite{Lazarow2017intro} is made to better understand the overall objective function within INN.
\item By adopting the Wasserstein distance into INN, we are able to generate images using a single CNN in WINN with even higher quality than those by INN that uses 20 CNNs (as seen in Figure \ref{fig:texture_big}, \ref{fig:texture_small}, \ref{fig:celebA-compare}, \ref{fig:svhn-compare}, and \ref{fig:cifar-compare}; the similar underlying CNN architectures are used in WINN and INN). WINN achieves a significant reduction in model complexity over INN, making the generator more practical.
\item Within texture modeling, INN and WINN are able to inherently model the input image space, making the synthesis of large texture images realistic, whereas GAN projects a noise vector onto the image space
making the image patch stitching more difficult (although extensions exist), as demonstrated in Figure \ref{fig:texture_big}.
\item To compare with the family of GAN models, we compute Inception scores using the standard procedure on the CIFAR-10 datasets and observed modest results. Here, we typically train 4-5 cascades
to boost the numbers but WINN with one CNN is already promising. Overall, modern GAN variants (e.g., \cite{WGAN-GP}) still outperform our WINN with better quality images. Some results are shown in Figure \ref{fig:cifar-compare}.
\item To test the robustness of the discriminative abilities of WINN, we directly make WINN into a discriminative classifier by training it on the standard MNIST and SVHN datasets. Not only are we able to improve over the previous ICN \cite{Jin2017intro} classifier for supervised classification, we also observe a large improvement in robustness against adversarial examples compared with the baseline CNN, ResNet, and the competing ICN.
\end{itemize}
}

In terms of other related work, we briefly discuss some existing methods below.

\noindent \textbf{Wasserstein GAN.} A closely related work to our WINN algorithm is the Wasserstein generative adversarial networks (WGAN) method \cite{WGAN,WGAN-GP}. While WINN adopts the Wasserstein distance as motivated by WGAN, our overall algorithm is still within the family of introspective neural networks (INN) \cite{Lazarow2017intro,Jin2017intro}. WGAN on the other hand is a variant of GAN with an improvement over GAN by having an objective that is easier to train. The level of difference between WINN and WGAN is similar to that between INN \cite{Lazarow2017intro,Jin2017intro} and GAN \cite{GAN}. The overall comparisons between INN and GAN have been described in \cite{Lazarow2017intro,Jin2017intro}.

\noindent \textbf{Generative ConvNets.} Recently, there has also been a cluster of algorithms developed in \cite{xie2016cooperative,xie2016theory,han2016learning}  where Langevin dynamics are adopted in generator CNNs. However, the models proposed in \cite{xie2016cooperative,xie2016theory,han2016learning} do not perform introspection (Figure \ref{fig:INN_pipeline}) and their generator and discriminator components are still somewhat separated; thus, their generators are not used as effective discriminative classifiers to perform state-of-the-art classification on standard supervised machine learning tasks. Their training processes are also more complex than those of INN and WINN.

\noindent \textbf{Deep energy models (DEMs) \cite{DEM}.}
DEM \cite{DEM} extends the standard density estimation by using multi-layer neural networks (MLNN) with a rather complex training procedure.
The probability model in DEM includes both the raw input and the features computed by MLNN. WINN instead takes a more general and simplistic form and is easier to train (see Eq. (\ref{eq:INN})). 
In general, DEM belongs to the minimum description length (MDL) family models in which the maximum likelihood is achieved.
WINN, instead, has a formulation being simultaneously discriminative and generative.

\section{Introspective Neural Networks}
\label{inn-intro}
\subsection{Brief introduction of INN}
We first briefly introduce the introspective neural network method (INNg) \cite{Lazarow2017intro} for generative modeling and its companion model \cite{Jin2017intro} which focuses on the classification aspect. The main motivation behind the INN work \cite{Lazarow2017intro,Jin2017intro} is to make a convolutional neural network classifier simultaneously discriminative and generative. A single CNN classifier is trained in an introspective manner to improve the standard supervised classification result \cite{Jin2017intro}, however, a sequence of CNNs (typically $10-60$) is needed to be able to synthesize images of good quality \cite{Lazarow2017intro}.

\begin{figure*}[!htp]
\begin{center}
\begin{tabular} {c}
\hspace{-5mm} \includegraphics[width=1.0\textwidth]{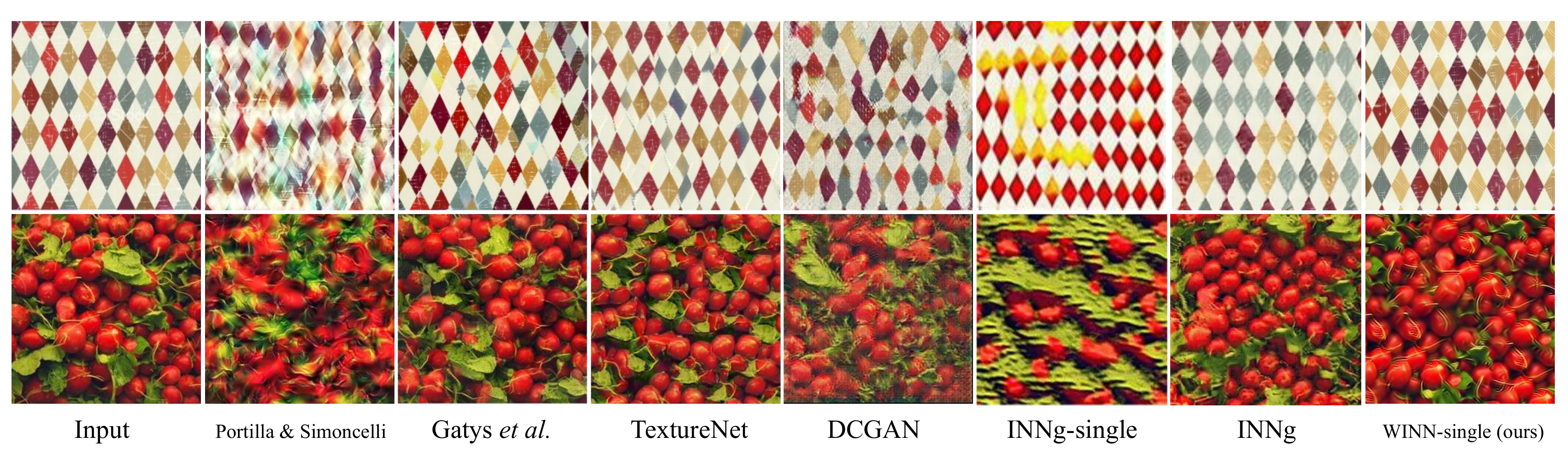}
\end{tabular}
\end{center}
\vspace{-6mm}
\caption{\footnotesize Comparison of texture synthesis algorithms. Gatys \textit{et al}. \cite{gatys2015texture}, and TextureNet \cite{ulyanov2016texture} results are from \cite{ulyanov2016texture}.}
\label{fig:texture_big}
\vspace{-2mm}
\end{figure*}

Figure \ref{fig:INN_pipeline} shows a brief illustration of a single introspective CNN classifier \cite{Jin2017intro}. We discuss our basic unsupervised formulation next. Suppose we are given a set of training examples: $S=\{\x_i \mid i=1,\ldots,n\}$ where we assume each $x_i \in \mathbb{R}^m$ --- e.g., $m = 4096$ for images of size $64 \times 64$. These will constitute positive examples of the patterns/targets we wish to model. The main idea of INN is to define pseudo-negative examples that are to be self-generated by the discriminative classifier itself. We therefore define label $y$ for each example $\x$, $y =+1$ if $\x$ is from the given training set and $y=-1$ if $\x$ is self-generated. Motivated by the generative via discriminative learning (GDL) framework \cite{tu2007learning}, one could try to learn the generative model for the given examples, $p(\x|y=+1)$, by a sequentially learned distribution of the {\bf pseudo-negative} samples,  $p_t(\x|y=-1; \W_t)$ which is abbreviated as $p_{\W_t}^-(\x)$ where $\W_t$ includes all the model parameters learned at step $t$. 
\begin{equation}
 p_{\W_t}^-(\x) =  \frac{1}{Z_{t}} \exp\{\w_t^{(1)} \cdot \phi(\x;\w_t^{(0)})\} \cdot p_{0}^-(\x), \; t=1,\ldots,T
\label{eq:INN}
\end{equation}
where $Z_t=\int \exp\{\w_t^{(1)} \cdot \phi(\x;\w_t^{(0)})\} \cdot p^-_{0}(\x) d\x$ and the initial distribution $p_0^-(\x)$ such as a Gaussian distribution over the entire space of $\x \in \mathbb{R}^m$. The discriminative classifier is a convolutional neural network (CNN) parameterized by $\W_t=(\w_t^{(0)}, \w_t^{(1)})$ where $\w_t^{(1)}$ denotes the weights of the top layer combining the features through $\phi(\x;\w_t^{(0)})$ (e.g., softmax layer) and $\w_t^{(0)}$ parameterizing the internal representations. The synthesis process through which pseudo-negative samples are generated is carried out by stochastic gradient Langevin dynamics \cite{welling2011bayesian} as
\[
\Delta \x = \frac{\epsilon}{2} \; \nabla (\w_t^{(1)} \cdot \phi(\x;\w_t^{(0)})) + \eta
\]
where $\eta \sim \mathcal{N}(0, \epsilon)$ is a Gaussian distribution and $\epsilon$ is the step size that is annealed in the sampling process. Overall, we desire
\begin{equation}
p_{\W_t}^-(\x) \stackrel{t=\infty}{\rightarrow} p(\x|y=+1),
\label{eq:pt_def}
\end{equation}
using the iterative reclassification-by-synthesis process \cite{Jin2017intro,Lazarow2017intro} guided by Eq. (\ref{eq:INN}).

\subsection{Connection to the Wasserstein distance}
\label{connection}

The overall training process, reclassification-by-synthesis, is carried out iteratively without an explicit objective function. The generative adversarial network (GAN) model \cite{goodfellow2014generative} instead has an objective function formulated in a minimax fashion with the generator and discriminator competing against each other. The Wasserstein generative adversarial network (WGAN) work \cite{WGAN} improves GAN \cite{goodfellow2014generative} by replacing the Jensen-Shannon distance with an efficient approximation of the Earth-Mover distance \cite{WGAN}. Also, there has been further generalization of the GAN family models in \cite{liu2017approximation}.

Let $p^+(\x) \equiv p(\x|y=+1)$ be the target distribution and 
$p_{\W}^-(\x) \equiv p(\x|y=-1; \W)$ be the pseudo-negative distribution parameterized by $\W$. Next, we show a connection between the INN framework and 
the WGAN formulation \cite{WGAN}, whose objective (rewritten with our notations) can be defined as
\begin{equation}
    \min_{\W} \max_{||f||_L \le 1} E_{\x \sim p^+}[f(\x)] - E_{\x \sim p_{\W}^-}[f(\x)],
\label{eq:WGAN}
\end{equation}
where $||f||_L \le 1$ denotes the space of 1-Lipschitz functions.
To build the connection between Eq. (\ref{eq:WGAN}) of WGAN and Eq. (\ref{eq:INN}) of INN, we first present the following lemma.

{\lemma \label{lm:1}
Considering $f(\x) = \ln \frac{p^+(\x)}{p_{\W}^-(\x)}$ and assuming its 1-Lipschitz property, we have a lower bound on the Wasserstein distance by
\begin{eqnarray}
& &\max_{||f||_L \le 1} E_{\x \sim p^+}[f(\x)] - E_{\x \sim p_{\W}^-}[f(\x)] \nonumber \\
&\geq&  KL(p^+||p_{\W}^-) + KL(p_{\W}^-||p^+)
\label{eq:JD}
\end{eqnarray}
where $KL(p||q)$ denotes the Kullback-Leibler divergence between the two distributions $p$ and $q$, and $KL(p^+||p_{\W}^-) + KL(p_{\W}^-||p^+)$ is the Jeffreys divergence.
}\\
\textbf{Proof:} see Appendix A.\\
Note that using the Bayes' rule, the ratio of the generative probabilities $\frac{p(\x|y=+1)}{p(\x|y=-1)}$ in Lemma \ref{lm:1} can be turned into the ratio of the discriminative probabilities $\frac{p(y=+1|\x)}{p(y=-1|\x)}$ assuming equal priors $p(y=+1)=p(y=-1)$.
{\theor \label{th:1} The introspective neural network formulation (Eq. (\ref{eq:INN})) implicitly minimizes a lower bound of the WGAN objective \cite{WGAN} (Eq. (\ref{eq:WGAN})).}\\
\textbf{Proof:} see Appendix B.

\section{Wasserstein Introspective Networks}
\begin{algorithm}[!htp]
\caption{\small Outline of WINN-single training algorithm. We use $k=3$ and $\lambda=10$.}
\label{al:INNg-cascade}
\begin{algorithmic}
{
{\small
\While {$\W_t$ has not converged}
    \State // Classification-step:
    \For {$k$ steps}
        \State Sample $m$ positive samples ${\scriptstyle \{\x_1^{+},\cdots, \x_m^{+}\}}$ from $S_+$.
        \State Sample $m$ {\footnotesize pseudo-negative samples} ${\scriptstyle \{\x_1^{-},\cdots, \x_m^{-}\}}$ from $S^t_{-}$.
        \State Sample $m$ random numbers ${\scriptstyle \{\alpha_1,\cdots, \alpha_m\}}$ from $U[0, 1]$.
        \State $\hat{\x}_i\leftarrow \alpha_i \x_i^{+}+(1-\alpha_i) \x_i^{-}$, $i=1,\ldots,m$.
        \State Perform stochastic gradient descent:
        \State ${\scriptstyle \nabla_{\W_t}\frac{1}{m}\sum_{i=1}^{m}\{[f_{\W_t}(\x_i^{-})-f_{\W_t}(\x_i^{+})]+ \lambda(\lVert \nabla_{\hat{\x}_i}f_{\W_t}(\hat{\x}_i)\rVert_2-1)^2\}}$.
    \EndFor
    \State // Synthesis-step:
    \State Sample $r$ noise samples $\{\x_1,\cdots, \x_{r}\}$ from $p_{0}^-(\x)$.
    \State Perform stochastic gradient ascent with early-stopping.
    \State $S^{t+1}_{-} \leftarrow S^{t}_{-}\cup\{\x_1,\cdots, \x_{r}\}$.
    \State $t \leftarrow t+1$.
\EndWhile
}
}
\end{algorithmic}
\end{algorithm}


Here we present the formulation for WINN building upon the formulation of the prior introspective learning works presented in Section \ref{inn-intro}.
\subsection{WINN algorithm}
We denote our unlabeled input training data as $S_{+}=\{\x_i|y_i=+1,i=1,\ldots,n\}$. Also, we denote the set of all the self-generated pseudo-negative samples up to step $t$ as $S_{-}^t=\{\x_i|y_i=-1, i=1,\ldots,l \}$. In other words, $S_{-}^t$ consists of pseudo-negatives $\x$ sampled from our model $p_{\W_t}^-(\x)$ for $t\ge 1$ where $\W_t$ is the model parameter vector at step $t$.
\vspace{2mm}

\noindent \textbf{Classification-step.}
The classification-step can be viewed as training a classifier to approximate the Wasserstein distance between $S_+$ and $S_{-}^t$ for $t\ge 1$. Note that we also keep pseudo-negatives from earlier stages -- which are essentially the mistakes of the earlier stages -- to prevent the classifier forgetting what it has learned in previous stages. We use CNNs parametrized by $\W_t$ as base classifiers. Let $f_{\W_t}(\cdot)$ denote the output of final fully connected layer (without passing through sigmoid nonlinearity) of the CNN. In the previous introspective learning frameworks \cite{Lazarow2017intro, Jin2017intro}, the classifier learning objective was to minimize the following standard cross-entropy loss function on $S_+ \cup S_{-}^t$:
\begin{equation}
{\scriptstyle  \mathcal{L}(\W_t) = - \{\mathbb{E}_{\x^+\sim p^+}\ln \sigma[+f_{\W_t}(\x^+)] +\mathbb{E}_{\x^-\sim p_{\W_t}^{-}} \ln \sigma[-f_{\W_t}(\x^-)] \} }\nonumber \\
\label{eq:loss_binary}
\vspace{-1mm}
\end{equation}

\noindent where $\sigma(\cdot)$ denotes the sigmoid nonlinearity.
Motivated by Section \ref{connection}, in WINN training we wish to minimize the following Wasserstein loss function by the stochastic gradient descent algorithm via backpropagation:

\begin{equation}
\vspace{0mm}
{\scriptstyle  \mathcal{L}(\W_t) = - \big[\mathbb{E}_{\x^+\sim p^+}f_{\W_t}(\x^+) -\mathbb{E}_{\x^-\sim p_{\W_t}^{-}} f_{\W_t}(\x^-) \big] } \\ 
\label{eq:loss_wasserstein}
\end{equation}

To enforce the function $f_{\W_t}$ to be $1$-Lipschitz, we add the following gradient penalty term \cite{WGAN-GP} to $\mathcal{L}(\W_t)$:

{\footnotesize
\begin{equation}
  \lambda \mathbb{E}_{\hat{\x}\sim p_{\hat{\x}}}[(\lVert \nabla_{\hat{\x}}f_{\W_t}(\hat{\x})\rVert_2-1)^2]\nonumber
\label{eq:loss_wasserstein2}
\end{equation}
}
where $\hat{\x}=\alpha \x^+ + (1-\alpha) {\x}^-$, $\x^+\sim p^+$, ${\x}^-\sim p^-_{\W_t}$, and $\alpha \sim U[0, 1]$.
\vspace{2mm}

\noindent \textbf{Synthesis-step.} Obtaining increasingly difficult pseudo-negative samples is an integral part of the introspective learning framework, as it is crucial for tightening the decision boundary. To this end, we develop an efficient sampling procedure under the Wasserstein formulation. After the classification-step, we obtain the following distribution of pseudo-negatives:
\begin{equation}
 p_{\W_t}^-(\x) =  \frac{1}{Z_{t}} \exp\{f_{\W_t}(\x)\} \cdot p_{0}^-(\x), \; t=1,\ldots,T
\label{eq:INN2}
\end{equation}
where $Z_t=\int \exp\{f_{\W_t}(\x)\} \cdot p^-_{0}(\x) d\x$; the initial distribution $p_0^-(\x)$ is a Gaussian distribution $G(\x; 0, \sigma^2)$ or the distribution defined in Appendix \textbf{D}. We find that the distribution of Appendix \textbf{D} encourages the diversity of sampled images.\\
The following equivalence is shown in \cite{Lazarow2017intro, Jin2017intro}:
\begin{equation}
  \frac{p (y=+1|\x;\W_t)}{p(y=-1|\x;\W_t)}=\exp\{f_{\W_t}(\x)\}.
\label{eq:INN-ratio}
\end{equation}
The sampling strategy of \cite{Lazarow2017intro, Jin2017intro} was to carry out gradient ascent on the term $\ln \frac{p (y=+1|\x;\W_t)}{p(y=-1|\x;\W_t)}$. In Lemma \ref{lm:1} we chose $f(\x)$ to be $\ln \frac{p^+(\x)}{p_{\W}^-(\x)}$. Using Bayes' rule, it is easy to see that $\nabla \ln \frac{p (y=+1|\x;\W_t)}{p(y=-1|\x;\W_t)}$ is loosely connected to $\nabla \ln \frac{p^+(\x)}{p_{\W_t}^-(\x)}$. Also, \cite{WGAN, WGAN-GP} argue that $f_{\W_t}(\x)$ correlates with the quality of the sample $\x$. This motivates us to use the following sampling strategy.
After initializing $\x$ by drawing a fair sample from $p_{0}^-(\x)$, we increase $f_{\W_t}(\x)$ using gradient ascent on the image $\x$ via backpropagation. Specifically, as shown in \cite{welling2011bayesian}, we can obtain fair samples from the distribution $p_{\W_t}^-$ using the following update rule:
\[
\Delta \x = \frac{\epsilon}{2} \; \nabla f_{\W_t}(\x) + \eta
\]
where $\epsilon$ is a time-varying step size and $\eta$ is a random variable following the Gaussian distribution $N(0,\epsilon)$. Gaussian noise term is added to make samples cover the full distribution. Inspired by \cite{isola2017image}, we found that injecting noise in the image space could be substituted by applying Dropout to the higher layers of CNN. In practice, we were able obtain the samples of enough diversity without step size annealing and noise injection.\\
As an early stopping criterion, we empirically find that the following is effective: (1) we measure the minimum and maximum $f_{\W_t}(\cdot)$ of positive examples; (2) we set the early stopping threshold to a random number from the uniform distribution between these two numbers. Intuitively, by matching the value of $f_{\W_t}(\cdot)$ positives and  pseudo-negatives, we expect to obtain pseudo-negative samples that match the quality of positive samples.

\subsection{Expanding model capacity} In practice, we find that the version with the single classifier -- which we call WINN-single -- is expressive enough to capture the generative distribution under variety of applications. The introspective learning formulation \cite{tu2007learning, Lazarow2017intro, Jin2017intro} allows us to model more complex distributions by adding a sequence of cascaded classifiers parameterized by $({\W^1},\ldots,{\W^K})$. Then, we can model the distribution as:
\begin{equation}
 p_{\W^k}^-(\x) = \frac{1}{Z_{t}} \exp\{f_{\W^k}(\x)\} \cdot  p_{\W^{k-1}}^-(\x),\;k=2,\ldots,K
\label{eq:INN-cascaded}
\end{equation}
In the next sections, we demonstrate the modeling capability of WINN under cascaded classifiers, as well as its agnosticy to the type of base classifier.

\subsection{GAN's discriminator vs. WINN's classifier}

\begin{figure}[!htb]
\vspace{-3mm}
\begin{center}
\begin{subfigure}{0.23\textwidth}
\centering
\includegraphics[width=0.8\textwidth]{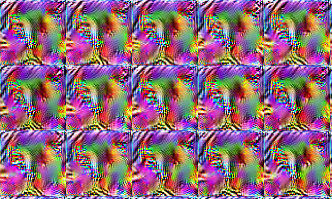}
\caption{WGAN-GP discriminator}
\end{subfigure}
\begin{subfigure}{0.23\textwidth}
\centering
\includegraphics[width=0.8\textwidth]{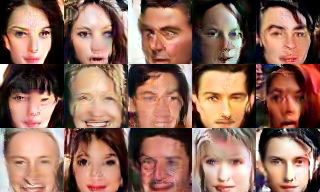}
\caption{WINN-single}
\end{subfigure}
\end{center}
\vspace{-5mm}
\caption{\footnotesize Synthesized images from the discriminators of WGAN and WINN trained on the CelebA dataset. Both use the same ResNet architecture for the discriminator as the one adopted in \cite{WGAN-GP}.}
\label{fig:discriminator-compare}
\vspace{-4mm}
\end{figure}

GAN uses the competing discriminator and the generator whereas WINN maintains a single model being both generative and discriminative. Some general comparisons between GAN and INN have been provided in \cite{Lazarow2017intro,Jin2017intro}. Below we make a few additional observations that are worth future exploration and discussions. 

{\small
\begin{itemize}
 \setlength\itemsep{0mm}
 \setlength{\itemindent}{0mm}
 \item First, the generator of GAN is a cost-effective option for image patch synthesis, as it works in a feed-forward fashion. However the generator of GAN is not meant to be trained as a classifier to perform the standard classification task, while the generator in the introspective framework is also a strong classifier. Section \ref{sec:mnist-classification} shows WINN to have significant robustness to external adversarial examples.
\item Second, the discriminator in GAN is meant to be a critic but not a generator. To show whether or not the discriminator in GAN can also be used as a generator, we train WGAN-GP \cite{WGAN-GP} on the CelebA face dataset. Using the same CNN architecture (ResNet from \cite{WGAN-GP}) that was used as GAN's discriminator, we also train a WINN-single model, making GAN's discriminator and WINN-single to have the identical CNN architecture. Applying the sampling strategy to WGAN-GP's discriminator allows us to synthesize image form WGAN-GP's discriminator as well and we show some samples in Figure \ref{fig:discriminator-compare} (a). These synthesized images are not like faces, yet they have been classified by the discriminator of WGAN-GP as ``real'' faces; this demonstrates the separation between the generator and the discriminator in GAN. In contrast, images synthesized by WINN-single's CNN classifier are faces like, as shown in Figure \ref{fig:discriminator-compare} (b). 
\item Third, the discriminator of GAN may not be used as a direct discriminative classifier for the standard supervised learning task. As shown and discussed in ICN \cite{Jin2017intro}, the introspective framework has the ability of classification for discriminator.
\end{itemize}
}


\section{Experiments}
\subsection{Implementation}
\label{winn_implementation}
\noindent \textbf{Classification-step.} For training the discriminator network, we use Adam \cite{kingma2014adam} with a mini-batch size of 100. The learning rate was set to 0.0001. We set $\beta_1=0.0,$ and $\beta_2=0.9$, inspired by \cite{WGAN-GP}. Each batch consists of 50 positive images sampled from the set of positives $S_+$ and 50 pseudo-negative images sampled from the set of pseudo-negatives $S_-$. In each iteration, we limit total number of training images to $10,000$.\\
\textbf{Synthesis-step.} For synthesizing pseudo-negative images via back-propagation, we perform gradient ascent on the image space. In the first cascade, each image is initialized with a noise sampled from the distribution described in Appendix \textbf{D}. In the later cascades, images are initialized with the images sampled from the last cascade. We use Adam with a mini-batch size of 100. The learning rate was set to 0.01. We set $\beta_1=0.9,$ and $\beta_2=0.99$.

\vspace{-2mm}
\subsection{Texture modeling}
\vspace{-2mm}

\begin{figure}[!htp]
\vspace{-2mm}
\begin{center}
\begin{tabular} {c}
\hspace{-5mm} \includegraphics[width=0.5\textwidth]{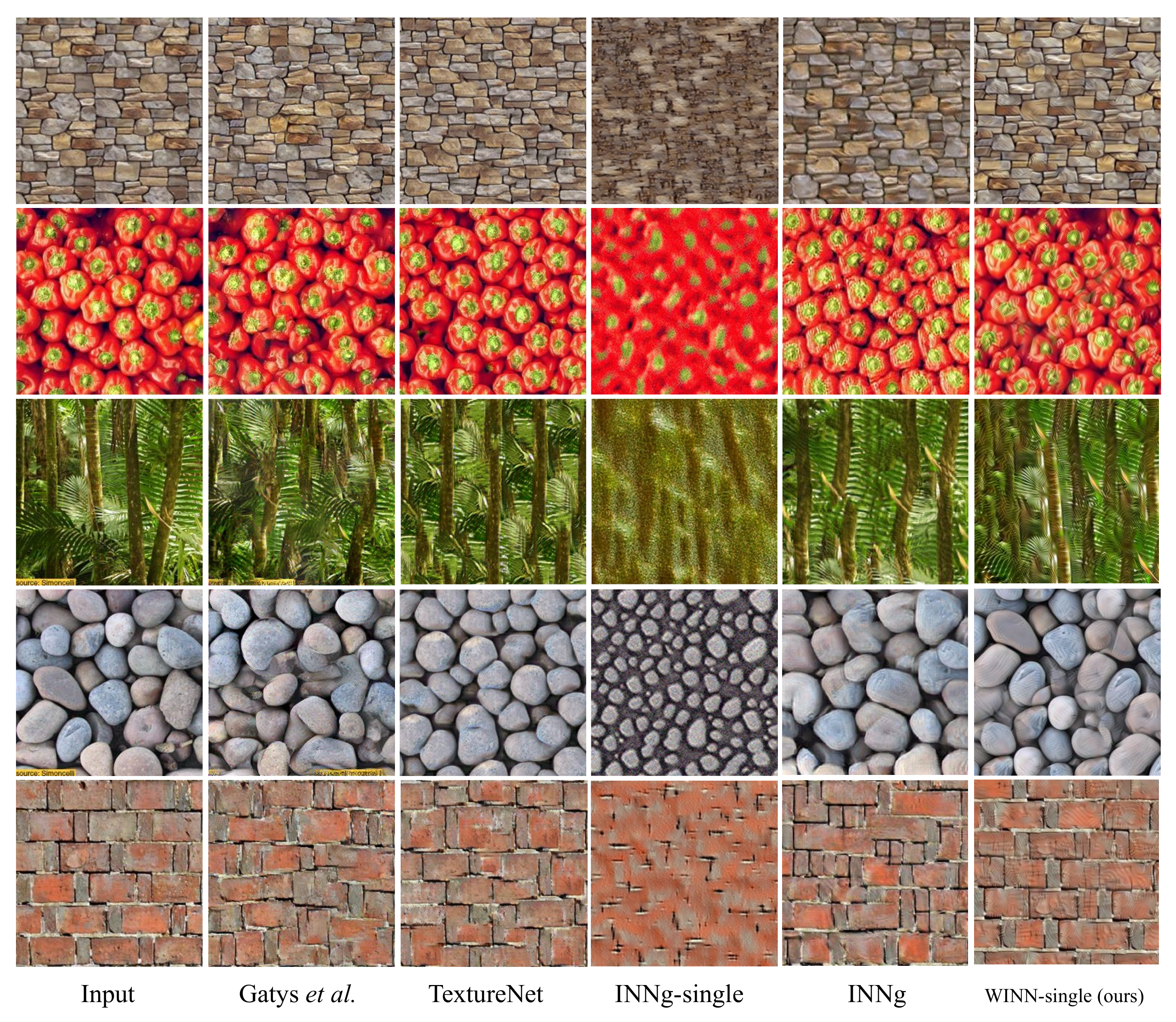}
\end{tabular}
\end{center}
\vspace{-8mm}
\caption{\footnotesize More texture synthesis results. Gatys \textit{et al}. \cite{gatys2015texture} and TextureNets \cite{ulyanov2016texture} results are from \cite{ulyanov2016texture}.}
\label{fig:texture_small}
\vspace{-4mm}
\end{figure}

We evaluate the texture modeling capability of WINN. For a fair comparison, we use the same 7 texture images presented in \cite {gatys2015texture} where each texture image has a size of 256 $\times$ 256. We follow the training method of Section \ref{winn_implementation} except that positive images are constructed by cropping 64 $\times$ 64 patches from the source texture image at random positions. We use network architecture of Appendix \textbf{C}. 
After training is done on the 64$\times$64 patch-based model, we try to synthesize texture images of arbitrary size using the anysize-image-generation method following \cite{Lazarow2017intro}. During the synthesis process, we keep a single working image of size 320$\times$320. Note that we expand the image so that center 256$\times$256 pixels are covered with equal probability. In each iteration, we sample 200 patches from the working image, and perform gradient ascent on the chosen patches. For the overlapping pixels between patches, we take the average of the gradients assigned to such pixels. We show synthesized texture images in Figure \ref{fig:texture_big} and \ref{fig:texture_small}. WINN-single shows a significant improvement over INNg-single and comparable results to INNg (using 20 CNNs). It is worth noting that \cite{gatys2015neural, ulyanov2016texture} leverage rich features of VGG-19 network pretrained on ImageNet. WINN and INNg instead train networks from scratch.

\vspace{-1mm}
\subsection{CelebA face modeling}
\vspace{-2mm}

\begin{figure}[!htb]
\vspace{-3mm}
\begin{center}
\begin{tabular} {ccc}
\hspace{-3mm}    \includegraphics [width=0.14\textwidth] {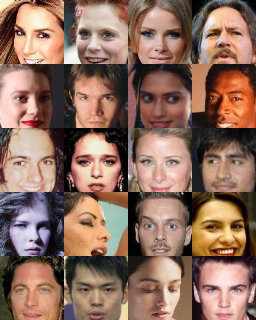} &
\hspace{-3mm}    \includegraphics [width=0.14\textwidth] {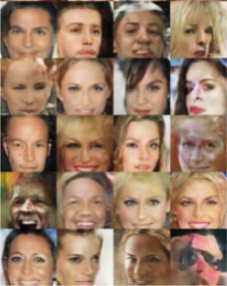} &
\hspace{-3mm}    \includegraphics [width=0.14\textwidth] {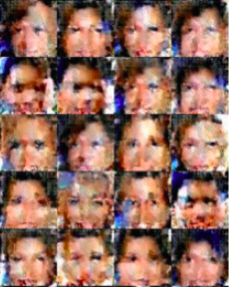} \\
\hspace{-3mm}	\footnotesize Real data & \hspace{-4mm} \footnotesize DCGAN & \hspace{-3mm} \footnotesize INNg-single \\
\hspace{-3mm}    \includegraphics [width=0.14\textwidth] {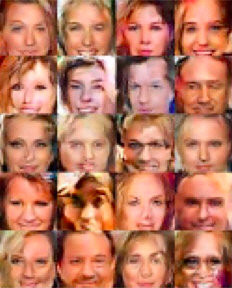} &
\hspace{-3mm}    \includegraphics [width=0.14\textwidth] {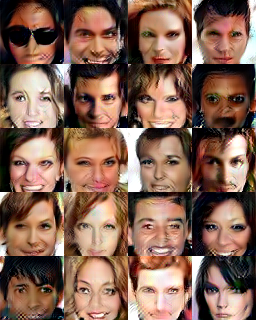} &
 \hspace{-3mm}   \includegraphics [width=0.14\textwidth] {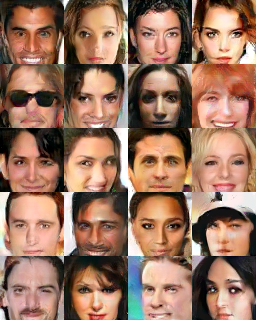}
 \\
\hspace{-3mm} \footnotesize INNg & \hspace{-3mm} \footnotesize WINN-single (ours) & \hspace{-2mm} \footnotesize WINN-4CNNs (ours)
\end{tabular}
\end{center}
\vspace{-7mm}
\caption{\footnotesize Images generated by various models trained on CelebA.}
\vspace{-3mm}
\label{fig:celebA-compare}
\end{figure}

The CelebA dataset \cite{CelebA} consists of $202,599$ face images of celebrities. This dataset has been widely used in the previous generative modeling works since it contains large pose variations and background clutters. The network architecture adopted here is described in Appendix \textbf{C}. In Figure \ref{fig:celebA-compare}, we show some synthesized face images using WINN-single and WINN, as well as those by DCGAN \cite{radford2015unsupervised}, INNg-single, and INNg \cite{Lazarow2017intro}. WINN-single attains  image quality even higher than that of INNg (12 CNNs).
\vspace{-2mm}
\subsection{SVHN modeling}
\vspace{-1mm}

\label{sec:svhn_unsupervised}


\begin{figure}[!htp]
\vspace{-3mm}
\begin{center}
\begin{subfigure}{0.14\textwidth}
\centering
\includegraphics[width=0.75\textwidth]{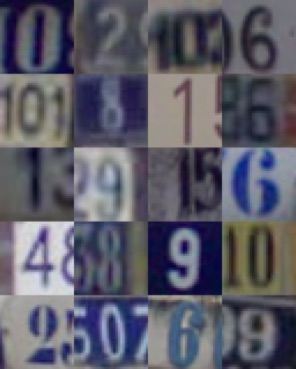}
\vspace{-1mm}
\caption*{\footnotesize Real data}
\vspace{1mm}
\end{subfigure}
\hspace{-3mm} 
\begin{subfigure}{0.14\textwidth}
\centering
\includegraphics[width=0.75\textwidth]{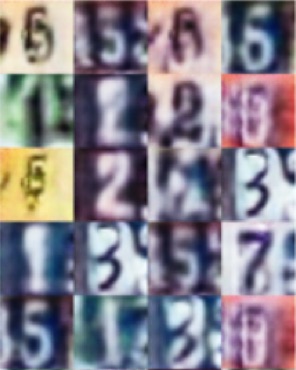}
\vspace{-1mm}
\caption*{\footnotesize DCGAN}
\vspace{1mm}
\end{subfigure}
\hspace{-3mm}
\begin{subfigure}{0.14\textwidth}
\centering
\includegraphics[width=0.75\textwidth]{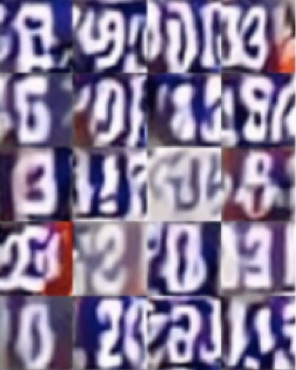}
\vspace{-1mm}
\caption*{\footnotesize INNg-single}
\vspace{1mm}
\end{subfigure}
\hspace{-3mm}
\begin{subfigure}{0.14\textwidth}
\centering
\includegraphics[width=0.75\textwidth]{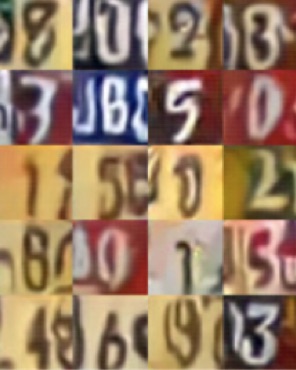}
\vspace{-1mm}
\caption*{\footnotesize INNg}
\end{subfigure}
\hspace{-3mm}
\begin{subfigure}{0.14\textwidth}
\centering
\includegraphics[width=0.75\textwidth]{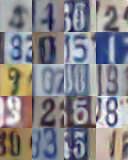}
\vspace{-1mm}
\caption*{\scriptsize WINN-single (ours)}
\end{subfigure}
\hspace{-3mm}
\begin{subfigure}{0.14\textwidth}
\centering
\includegraphics[width=0.75\textwidth]{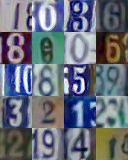}
\vspace{-1mm}
\caption*{\scriptsize WINN-4CNNs (ours)}
\end{subfigure}
\end{center}
\vspace{-7mm}
\caption{\footnotesize Images generated by various models trained on SVHN. DCGAN \cite{radford2015unsupervised} result is from \cite{Lazarow2017intro}.}
\label{fig:svhn-compare}
\vspace{-2mm}
\end{figure}


SVHN \cite{SVHN} consists of $32 \times 32$ images from Google Street View. It contains $73,257$ training images, $26,032$ test images, and $531,131$ extra images. We use only the training images for the unsupervised SVHN modeling. We use the ResNet architecture described in \cite{WGAN-GP}. Generated images by WINN-single and WINN (4 CNN classifiers) as well as DCGAN and INN are shown in Figure \ref{fig:svhn-compare}. The improvement of WINN over INNg is evident.

\vspace{-2mm}
\subsection{CIFAR-10 modeling}
\label{sec:cifar_unsupervised}
\vspace{-2mm}
\begin{table}[H]
\vspace{-3mm}
\centering
\caption{\footnotesize Inception score on CIFAR-10. ``-L'' in Improved GANs means without labels.}
\vspace{-2mm}
\label{table:inception}
\scalebox{0.8}{
\begin{tabular}{l|l}
    Method & Score\\
    \Xhline{4\arrayrulewidth}
    Real data & $11.95 \pm .20$ \\
    \hline
    WGAN-GP \cite{WGAN-GP}   & ${\bf 7.86} \pm .07$ \\
    WGAN \cite{WGAN}         & $5.88 \pm .07$ \\
    DCGAN \cite{radford2015unsupervised} (in \cite{SGAN})    & $6.16 \pm .07$ \\
    ALI \cite{ALI} (in \cite{DFM})         & $5.34 \pm .05$ \\
    Improved GANs (-L) \cite{salimans2016improved} & $4.36 \pm .04$ \\
    \hline
    INNg-single \cite{Lazarow2017intro}  & $1.95 \pm .01$ \\
    INNg \cite{Lazarow2017intro}         & $3.04 \pm .02$ \\
    WINN-single (ours) & $4.62 \pm .05$ \\
    WINN-5CNNs (ours) & $5.58 \pm .05$
\end{tabular}
}
\vspace{-2mm}
\end{table}

\begin{figure}[!htb]
\vspace{-3mm}
\begin{center}
\begin{tabular}{cccc}
\hspace{-2mm} \includegraphics[width=0.11\textwidth]{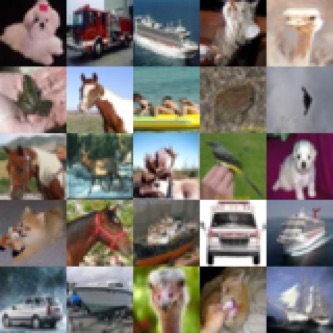} &
\hspace{-4mm} \includegraphics[width=0.11\textwidth]{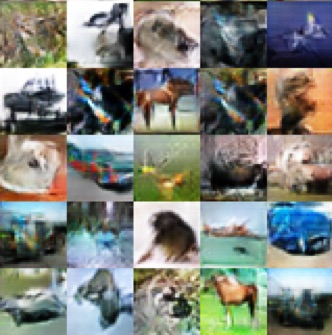} &
\hspace{-4mm} \includegraphics[width=0.11\textwidth]{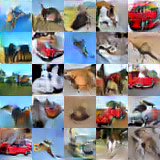} &
\hspace{-4mm} \includegraphics[width=0.11\textwidth]{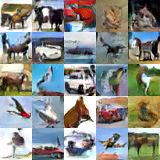} \\
{\scriptsize Real data} & \hspace{-2mm}{\scriptsize DCGAN} & \hspace{-4mm} {\scriptsize WINN-single (ours)} & \hspace{-3mm} {\tiny WINN-5CNNs (ours)}
\end{tabular}
\vspace{-3mm}
\caption{\footnotesize Images generated by models trained on CIFAR-10.}
\label{fig:cifar-compare}
\vspace{-6mm}
\end{center}
\end{figure}

CIFAR-10 \cite{cifar10} consists of $50,000$ training images and $10,000$ test images of size $32\times32$ in 10 classes. We use training images augmented by horizontal flips \cite{krizhevsky2012imagenet} for unsupervised CIFAR-10 modeling. We use the ResNet given in \cite{WGAN-GP}. Figure \ref{fig:cifar-compare} shows generated images by various models.

To measure the semantic discriminability, we compute the Inception scores \cite{salimans2016improved} on $50,000$ generated images.
WINN shows its clear advantage over INN. WINN-5CNNs produces a result close to WGAN but there is still a gap to the state-of-the-art results by WGAN-GP.


\vspace{-2mm}
\subsection{Image classification and adversarial examples \label{sec:mnist-classification}}
\vspace{-0mm}
To demonstrate the robustness of WINN as a discriminative classifier, we present experiments on the supervised classification tasks.
\noindent 
\begin{table}[!htp]
\vspace{-1mm}
\begin{center}
    \caption{\footnotesize Test errors on MNIST and SVHN. When training on SVHN, we only use training set. All results except WINN-single and baseline ResNet-32 are from \cite{Jin2017intro}. \cite{Jin2017intro} adopted DCGAN \cite{radford2015unsupervised} discriminator as their CNN architecture. The advantage of WINN over a vanilla CNN is evident. When applied to a stronger baseline such as ResNet-32, WINN is not losing ResNet's superior classification capability in standard supervised classification, while attaining special generative capability and robustness to adversarial attacks (see Table \ref{table:mnist-adversarial} and \ref{table:svhn-adversarial}) that do not exist in ResNet. The images below on the right are the generated samples by the corresponding WINN (ResNet-32) classifiers reported in the table.}    \label{tb:mnist-classification}
    \vspace{-2mm}
\begin{tabular}{cc}
\hspace{-2.5mm}
\scalebox{0.85}{
    \begin{tabular}{c|c}
        Method & Error\\
        \Xhline{4\arrayrulewidth}
        {\bf MNIST} & \\
        Baseline vanilla CNN (4 layers) & $0.89 \%$ \\
        CNN + GDL \cite{tu2007learning} & $0.85 \%$ \\
        CNN + DCGAN \cite{radford2015unsupervised} & $0.84 \%$ \\
        ICN \cite{Jin2017intro} & $0.81 \%$ \\
        WINN-single vanilla (ours) & $0.67\%$ \\
        \hline
        Baseline ResNet-32 & $\bf{0.45\%}$ \\
        WINN-single ResNet-32 (ours) & $0.48\%$ \\
        \Xhline{4\arrayrulewidth}
        {\bf SVHN} & \\
        Baseline ResNet-32 & $4.64\%$ \\
        WINN-single ResNet-32 (ours) & $\bf{4.50\%}$ \\
    \end{tabular}
} & 
\hspace{-5mm}

\begin{tabular} {c}

\begin{minipage}{.2\textwidth}
\vspace{22mm}
\includegraphics[width=0.6\linewidth]{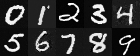}

\end{minipage}
\vspace{2mm}
\\
\begin{minipage}{.2\textwidth}
\includegraphics[width=0.6\linewidth]{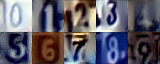}
\end{minipage}
\end{tabular}

\end{tabular}
\end{center}

\vspace{-3mm}
\end{table}

\vspace{-0mm}

\noindent \textbf{Training Methods.} We add the Wasserstein loss term to the ICN \cite{Jin2017intro} loss function, obtaining the following:
\vspace{-1mm}
\begin{eqnarray*}
 \mathcal{L}(\W_t) = &-  \sum_{\x_i \in S_+}^{} \ln \frac{\exp\{\w_t^{(1)_{y_i}} \cdot \phi(\x_i;\w_t^{(0)}) \} }{\sum_{k=1}^K \exp\{\w_t^{(1)_k} \cdot \phi(\x_i;\w_t^{(0)}) \} }  \\
 &+ \alpha \bigg( \sum_{\x_i \in S_{-}^t}^{} f_{\W_t}(\x_i) - \sum_{\x_i \in S_+}^{} f_{\W_t}(\x_i)   \\
 &+  \lambda \mathbb{E}_{\hat{\x}\sim p_{\hat{\x}}}[(\lVert \nabla_{\hat{\x}}f_{\W_t}(\hat{\x})\rVert_2-1)^2] \bigg)\nonumber
\end{eqnarray*}
\vspace{0mm}
where $\W_t=<\w_t^{(0)}, \w_t^{(1)_1},...,\w_t^{(1)_K}>$, $\w_t^{(0)}$ denotes the internal parameters for the CNN, and $\w_t^{(1)_k}$ denotes the top-layer weights for the $k$-th class. 
In the experiments, we set the weight of the WINN loss, $\alpha$, to 0.01. We use the vanilla network architecture resembling \cite{Jin2017intro} as the baseline CNN, which has less filters and parameters than the one in \cite{Jin2017intro}. We also use a ResNet-32 architecture with Layer Normalization \cite{lyr_norm} on MNIST and SVHN. In the classification-step, we use Adam with a fixed learning rate of 0.001, $\beta_1$ of $0.0$. In the synthesis-step, we use the Adam optimizer with a learning rate of $0.02$ and $\beta_1$ of $0.9$. Table \ref{tb:mnist-classification} shows the errors on MNIST and SVHN. 


\vspace{1mm}

\begin{table}[!htb]
\vspace{-3mm}
\caption{\footnotesize Adversarial examples comparison between the baseline CNN and WINN on MNIST. We first generate $N$ adversarial examples using method A and count the number of adversarial examples misclassified by A ($=N_A$). \textbf{Adversarial error} of A is defined as test error rate against adversarial examples ($=N_A/N$). Then among A's mistakes, we count the number of adversarial examples misclassified by B ($=N_{A\cap B}$). Then \textbf{error correction rate} by B is $1-N_{A\cap B}/N_A$. $\uparrow$ means higher the better; $\downarrow$ means lower the better. The comparisons are made in two groups: the first group builds on top a vanilla 4 layer CNN and the second group adopts ResNet-32. Note that the corresponding errors for these classifiers on the standard MNIST supervised classification task can be seen in Table \ref{tb:mnist-classification}.}
\label{table:mnist-adversarial}
\vspace{-6mm}
\begin{center}
\scalebox{0.65}{
\hspace{-6mm}
    \begin{tabular}{c|c|c|c}
         & Adversarial error &  Correction rate  & Correction rate \\
         \textbf{Method} & of \textbf{Method} $\downarrow$ & by \textbf{Method} $\uparrow$ & by \textbf{Baseline} $\downarrow$ \\
        \Xhline{4\arrayrulewidth}
        Baseline vanilla CNN & $32.41\%$ & - & - \\
        ICN \cite{Jin2017intro} & $19.02\%$  & $52.58\%$ & $ 58.68\%$\\
        WINN-single vanilla (ours) & ${\bf 7.99}\%$ & ${\bf 90.00}\%$ & ${\bf 46.93}\%$ \\
        \hline
        Baseline ResNet-32 & $11.28\%$ & - & - \\
        WINN-single ResNet-32 (ours) & ${\bf 2.05}\%$ & ${\bf 89.68}\%$ & $\bf 43.69\%$ \\
    \end{tabular}
}
\end{center}
\vspace{-8mm}
\end{table}
\noindent \textbf{Robustness to adversarial examples.}
It is argued in \cite{goodfellow2014generative} that discriminative CNN's vulnerability to adversarial examples primarily arises due to its linear nature. Since the reclassification-by-synthesis process helps tighten the decision boundary (Figure \ref{fig:INN_pipeline}), one might expect that CNNs trained with the WINN algorithm are more robust to adversarial examples. Note that unlike existing methods for adversarial defenses \cite{goodfellow2014explaining, FGSM2}, our method does not train networks with specific types of adversarial examples. With test images of MNIST and SVHN, we adopt ``fast gradient sign method'' \cite{goodfellow2014generative} ($\epsilon = 0.125$ for MNIST and $\epsilon = 0.005$ for SVHN) to generate adversarial examples clipped to range $[-1, 1]$, which differs from \cite{Jin2017intro}. We experiment with two networks having the same architecture and only differing in training method (the standard cross-entropy loss vs. the WINN procedure). We call the former as the baseline CNN. We summarize the results in Table \ref{table:mnist-adversarial}. Compared to ICN \cite{Jin2017intro}, WINN significantly reduces the adversarial error to 7.99\% and improves the correction rate to 90.00\%. In addition, we have adopted the ResNet-32 architecture into WINN. See Table \ref{table:mnist-adversarial} and \ref{table:svhn-adversarial}. We still obtain the adversarial error reduction and correction rate improvement on MNIST and SVHN ($\epsilon = 0.005$) with ResNet-32. Our observation is that WINN is not necessarily improving over a strong baseline for the supervised classification task but its advantage on adversarial attacks is evident.

\begin{table}[H]
\vspace{-2mm}
\caption{\footnotesize Adversarial examples comparison between the baseline ResNet-32 \cite{he2016deep} and WINN on SVHN. Note that the corresponding errors for these classifiers on the standard supervised SVHN classification task can be seen in Table \ref{tb:mnist-classification}.}
\label{table:svhn-adversarial}
\vspace{-6mm}
\begin{center}
\scalebox{0.63}{
\hspace{-4mm}
    \begin{tabular}{c|c|c|c}
         & Adversarial error &  Correction rate  & Correction rate \\
         \textbf{Method} & of \textbf{Method} &  by \textbf{Method} $\uparrow$ & by \textbf{Baseline} $\downarrow$ \\
        \Xhline{4\arrayrulewidth}
        Baseline ResNet-32 & $29.29\%$ & - & - \\
        WINN-single ResNet-32 (ours) & ${\bf 19.53}\%$ & $ 74.39\%$ & $51.37\%$ \\
    \end{tabular}
}
\end{center}
\vspace{-6mm}
\end{table}


\vspace{-1mm}
\subsection{Agnostic to different architectures}
\begin{figure}[!htb]
\begin{center}
\begin{tabular} {cc}
    \includegraphics [width=0.19\textwidth] {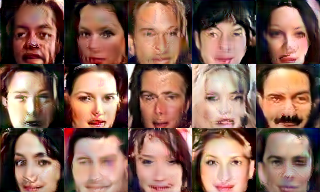} &
    \includegraphics [width=0.19\textwidth] {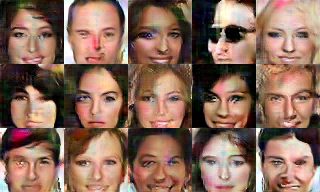} \\
	\footnotesize WINN-single (ResNet-13) & \footnotesize WINN-single (DenseNet-20) \\
\end{tabular}
\end{center}
\vspace{-7mm}
\caption{\footnotesize Synthesized CelebA images with varying network architectures. We use a single CNN for each experiment.}
\label{fig:celebA-agnos}
\vspace{-4mm}
\end{figure}

\vspace{-2mm}
In Figure \ref{fig:celebA-agnos}, we demonstrate our algorithm being agnostic to the type of classifier, by varying network architectures to ResNet \cite{he2016deep} and DenseNet \cite{huang2017dense}. Little modification was required to adapt two architectures for WINN.

\vspace{-3mm}
\section{Conclusion}
\vspace{-2mm}

In this paper, we have introduced Wasserstein introspective neural networks (WINN) that produce encouraging results as a generator and a discriminative classifier at the same time. WINN is able to achieve model size reduction over the previous introspective neural networks (INN) by a factor of $20$. In most of the images shown in the paper, we find a single CNN classifier in WINN being sufficient to produce visually appealing images as well as significant error reduction against adversarial examples. WINN is agnostic to the architecture design of CNN and we demonstrate results on three networks including a vanilla CNN, ResNet \cite{he2016deep}, and DenseNet \cite{huang2017dense} networks where popular CNN discriminative classifiers are turned into generative models under the WINN procedure. WINN can be adopted in a wide range of applications in computer vision such as image classification, recognition, and generation. \\
\noindent {\bf Acknowledgements}.
This work is supported by NSF IIS-1717431 and NSF IIS-1618477. The authors thank Justin Lazarow, Long Jin, Hubert Le, Ying Nian Wu, Max Welling, Richard Zemel, and Tong Zhang for valuable discussions.

\section{Appendix}

\noindent {\bf A. Proof of Lemma} \ref{lm:1}.

\textbf{Proof.} Plugging $f(\x) = \ln \frac{p^+(\x)}{p_{\W}^-(\x)}$ into Eq.  (\ref{eq:WGAN}), we have
\begin{eqnarray}
& & {\scriptstyle E_{\x \sim p^+}[f(\x)] - E_{\x \sim p_{\W}^-}[f(\x)]}  \nonumber \\
& = & {\scriptstyle E_{\x \sim p^+}[\ln \frac{p^+(\x)}{p_{\W}^-(\x)}] - E_{\x \sim p_{\W}^-}[\ln \frac{p^+(\x)}{p_{\W}^-(\x)}] } \nonumber \\
 &=&  {\scriptstyle \int p^+(\x) \ln \frac{p^+(\x)}{p_{\W}^-(\x)} d\x - \int p_{\W}^-(\x) \ln \frac{p^+(\x)}{p_{\W}^-(\x)} d\x } \nonumber \\
 &=&  {\scriptstyle KL(p^+||p_{\W}^-) + KL(p_{\W}^-||p^+) } \qquad \qquad \qquad \qquad \Box \nonumber
\end{eqnarray}

\noindent {\bf B. Proof of Theorem} \ref{th:1}.

{\corollary \label{cl:1}The Jeffreys divergence in Eq. (\ref{eq:JD}) of lemma \ref{lm:1} is lower and upper bounded by $KL(p^+||p_{\W}^-)$ up to some multiplicative constant

${\scriptstyle (1+\frac{p^+_{min}}{2}) KL(p^+||p_{\W}^-) \le JD(p^+;p_{\W}^-) \le (1+\frac{2}{p^+_{min}}) KL(p^+||p_{\W}^-) }$, where 
$JD(p^+;p_{\W}^-) = KL(p^+||p_{\W}^-) + KL(p_{\W}^-||p^+)$ is the Jeffreys divergence.}

\textbf{Proof.} Based on the Pinsker's inequality \cite{pinsker1960information,sason2016f}, it is observed that
\vspace{-1mm}
\[
KL(p_{\W}^-||p^+) \ge \frac{1}{2} |p_{\W}^--p^+|^2 \log e,
\]
\vspace{-1mm}
where $|p_{\W}^--p^+|$ is total variation (TV) distance.
From \cite{sason2016f} we also have
\vspace{-0mm}
\[
KL(p_{\W}^-||p^+) \le \frac{\log e}{p^+_{min}} |p_{\W}^--p^+|^2,
\]
\vspace{-0mm}
where $p^+_{min}=\min_{\x} p^+(\x)$.
Applying the above bounds to $KL(p^+ || p_{\W}^-)$ and using the symmetry of the TV distance $|p_{\W}^--p^+| \equiv |p^+ - p_{\W}^-|$,
\[
\frac{p^+_{min}}{2} KL(p^+||p_{\W}^-)  \le KL(p_{\W}^-||p^+) \le \frac{2}{p^+_{min}} KL(p^+||p_{\W}^-).
\]
Plugging the equation above into the Jeffreys divergence, we observe that $KL(p^+||p_{\W}^-) + KL(p_{\W}^-||p^+)$ is upper and lower bounded by by $KL(p^+||p_{\W}^-)$.
\hspace*{\fill} $\Box$


Now we can look at theorem \ref{th:1}. It was shown in \cite{Jin2017intro} that Eq. (\ref{eq:INN}) reduces $KL(p^+||p_{\W}^-)$, which bounds the Jeffreys divergence $KL(p^+||p_{\W}^-) + KL(p_{\W}^-||p^+)$ as shown in corollary \ref{cl:1}. Lemma \ref{lm:1} shows the connection between Jeffreys divergence and the WGAN objective (Eq. (\ref{eq:WGAN})) when $f(\x) = \ln \frac{p^+(\x)}{p_{\W}^-(\x)}$. We therefore can see that the formulation of introspective neural networks (Eq. (\ref{eq:INN})) connects to a lower bound of the WGAN \cite{WGAN} objective (Eq. (\ref{eq:WGAN})).
\hspace*{\fill} $\Box$

\vspace{1mm}
\noindent {\bf C. Texture and CelebA Modeling Architecture}. Inspired by \cite{NVIDIA}, we design a CNN architecture for 64 $\times$ 64 image  as in Table \ref{nVIDIA}. We use Swish \cite{swish} non-linearity after each convolutional layer. We add Layer Normalization \cite{lyr_norm} after each convolution except the first layer, following \cite{WGAN-GP}.
\vspace{-2mm}
\begin{table}[!htp]
\vspace{-2mm}
\centering
\caption{\footnotesize Network architecture for the texture and face image modeling.}
\vspace{-4mm}
\label{nVIDIA}
\begin{tabular}{cc}
\hspace{-1mm} {\scriptsize Textures and CelebA} & \hspace{-6mm} {\scriptsize Alternative Initialization} \\
\hspace{-1mm}\scalebox{0.5}{
\begin{tabular}{c|c|c}
Layer     & Filter size/stride & Output size \\
\Xhline{4\arrayrulewidth}
Input     &                          & 64$\times$64$\times$3     \\
\hline
Conv3-32  & 3$\times$3/1              & 64$\times$64$\times$32    \\
Conv3-64  & 3$\times$3/1              & 64$\times$64$\times$64    \\
Avg pool   & 2$\times$2/2             & 32$\times$32$\times$64   \\
\hline
Conv3-64  & 3$\times$3/1              & 32$\times$32$\times$64    \\
Conv3-128  & 3$\times$3/1             & 32$\times$32$\times$128  \\
Avg pool   & 2$\times$2/2             & 16$\times$16$\times$128  \\
\hline
Conv3-128  & 3$\times$3/1             & 16$\times$16$\times$128  \\
Conv3-256  & 3$\times$3/1             & 16$\times$16$\times$256  \\
Avg pool   & 2$\times$2/2             & 8$\times$8$\times$256    \\
\hline
Conv3-256  & 3$\times$3/1             & 8$\times$8$\times$256    \\
Conv3-512  & 3$\times$3/1             & 8$\times$8$\times$256    \\
Avg pool   & 2$\times$2/2             & 4$\times$4$\times$512    \\
\hline
FC-1    &                   & 1$\times$1$\times$1         \\
\end{tabular}
}
& \hspace{-3mm}
\scalebox{0.6}{
\hspace{-5mm} \begin{tabular}{c|c|c}
Layer     & Filter size/stride & Output size \\
\Xhline{4\arrayrulewidth}
Input     &                          & 4$\times$4$\times$512     \\
\hline
Conv5-256  & 5$\times$5/1             & 4$\times$4$\times$256    \\
\hline
Upsample  &                           & 8$\times$8$\times$256      \\
Conv5-128  & 5$\times$5/1             & 8$\times$8$\times$128    \\
\hline
Upsample  &                           & 16$\times$16$\times$128      \\
Conv5-64  & 5$\times$5/1             & 16$\times$16$\times$64      \\
\hline
Upsample  &                           & 32$\times$32$\times$64      \\
Conv5-64  & 5$\times$5/1             & 32$\times$32$\times$3      \\
\hline
Upsample  &                           & 64$\times$64$\times$3    
\end{tabular}
}
\end{tabular}
\vspace{-5mm}
\end{table}

\begin{wraptable}{r}{0.4\linewidth} \small
\vspace{-5mm}

\vspace{-5mm}
\end{wraptable}

\noindent {\bf D. Alternative Initializations}.
We sample an initial pseudo-negative image by applying an operation defined by the network above to a tensor of size $4\times4\times512$ sampled from $U[-1, 1]$. The weights of the network are sampled from $G(0, 0.1^2)$. We do not apply any nonlinearities in the network. We add Layer Normalization \cite{lyr_norm} after each convolution except the last layer. 

{\small
\bibliographystyle{ieee}
\bibliography{egbib}
}

\end{document}